\pdfoutput=1
\documentclass[11pt]{article}

\usepackage{acl}
\usepackage{times}
\usepackage{latexsym}
\usepackage{makecell}

\usepackage[T1]{fontenc}
\usepackage{pgfplots} 
\usepackage[utf8]{inputenc}
\usepackage{multirow}
\usepackage{amsmath}
\usepackage{capt-of}
\usepackage{tabularx}
\usepackage{epsfig}
\usepackage{amssymb}
\usepackage{amsfonts}
\usepackage{booktabs}
\usepackage{scalerel}
\usepackage[inline]{enumitem}
\usepackage{listings}
\usepackage{varwidth}
\usepackage[export]{adjustbox}
\usepackage{tikz}
\usetikzlibrary{tikzmark}
\usepackage{cleveref}

\usepackage{stmaryrd}
\usepackage{bbm}

\usepackage{algorithm}
\usepackage[noend]{algpseudocode}

\definecolor{deepblue}{rgb}{0,0,0.5}
\definecolor{officeblue}{RGB}{0,102,204}
\definecolor{deepred}{rgb}{0.6,0,0}
\definecolor{deepgreen}{rgb}{0,0.5,0}
\definecolor{mybrickred}{RGB}{182,50,28}

\definecolor{fillcolor}{RGB}{216,217,252}

\algnewcommand\algorithmicrequireb{{\hspace{0.85cm}}}
\algnewcommand\INPTDESCB{\item[\algorithmicrequireb]}

\algnewcommand\algorithmicfuncdesc{\textbf{Function:}}
\algnewcommand\FUNCDESC{\item[\algorithmicfuncdesc]}
\algnewcommand\algorithmicfuncdescb{{\hspace{1.48cm}}}
\algnewcommand\FUNCDESCB{\item[\algorithmicfuncdescb]}
\algnewcommand{\algorithmicgoto}{\textbf{goto}}
\algnewcommand{\Goto}[1]{\algorithmicgoto~\ref{#1}}

\usepackage{amsmath,amsfonts,bm}

\def\eqref#1{equation~\ref{#1}}
\def\1{\bm{1}}

\DeclareMathAlphabet{\mathsfit}{\encodingdefault}{\sfdefault}{m}{sl}
\SetMathAlphabet{\mathsfit}{bold}{\encodingdefault}{\sfdefault}{bx}{n}

\newcommand\ours{SciMRC}
\newcommand\modelT{PerT5}
\newcommand\modelLED{PerLED}

\newcommand\Qa{\textsc{QASPER}\space}

\title{\ours{}: Multi-perspective Scientific Machine Reading Comprehension}

\author{Xiao Zhang$^{123}$\thanks{\ \ Co-first authors with equal contributions.},~~Heqi Zheng$^{123}$\samethanks[1],~~Yuxiang Nie$^{123}$\samethanks[1],~~Heyan Huang$^{123}$,~~Xian-Ling Mao$^{123}$\\
$^{1}$School of Computer Science and Technology, Beijing Institute of Technology\\
$^2$Beijing Engineering Research Center of High Volume Language Information Processing\\
and Cloud Computing Applications\\
$^3$Southeast Academy of Information Technology, Beijing Institute of Technology\\
\texttt{\{yotta,heqizheng,yuxiangnie,hhy63,maoxl\}@bit.edu.cn}\\}

\newcommand*\samethanks[1][\value{footnote}]{\footnotemark[#1]}

\usepackage{pgfplots}
\pgfplotsset{compat=1.17}

\begin{document}
\maketitle
\begin{abstract}

Scientific machine reading comprehension (SMRC) aims to understand scientific texts through interactions with humans by given questions. As far as we know, there is only one dataset focused on exploring full-text scientific machine reading comprehension. However, the dataset has ignored the fact that different readers may have different levels of understanding of the text, and only includes single-perspective question-answer pairs, leading to a lack of consideration of different perspectives. To tackle the above problem, we propose a novel multi-perspective SMRC dataset, called \ours{}, which includes perspectives from beginners, students and experts. Our proposed \ours{} is constructed from 741 scientific papers and 6,057 question-answer pairs. Each perspective of beginners, students and experts contains 3,306, 1,800 and 951 QA pairs, respectively. The extensive experiments on \ours{} by utilizing pre-trained models suggest the importance of considering perspectives of SMRC, and demonstrate its challenging nature for machine comprehension.

% Our proposed dataset and experiments provide valuable insights and a new direction for future research in scientific MRC, in which we hope to improve the ability of machines to comprehend scientific texts from different perspectives.
\end{abstract}

\section{Introduction}
Scientific machine reading comprehension (SMRC) aims to understand scientific texts through interactions with humans by given questions. The ability of machines to understand and make sense of scientific texts is crucial for many applications such as scientific research \cite{Scicachola-etal-2020-tldr,Scibeltagy-etal-2019-scibert,Scimarie-etal-2021-scientific}, education \cite{SciChica2008PedagogicallyUEEducation,SciBianchi2019ScientificEAEducation} and industry \cite{SciZulfiqar2018IsMT,Sci-bruches-etal-2022-terminator,sci-erera-etal-2019-summarization}. With the increasing amount of scientific literature being produced, the need \cite{Sci-wadden-etal-2020-fact,Scisadat-caragea-2022-scinli,dasigi2021dataset} for machines to understand these texts is becoming more pressing. 

As far as we know, there is only one dataset \cite{dasigi2021dataset} focused on exploring full-text scientific machine reading comprehension, which is proposed to improve MRC models in seeking information from specific papers with questions. However, the dataset has ignored the fact that different readers may have different levels of understanding of the text,  and only includes single-perspective question-answer pairs from annotators whose background is NLP, which leads to a lack of consideration of different perspectives, especially for beginner's and expert's perspectives. Different perspectives correspond to different types of problems,  which requires different levels of understanding. It will help us analyze and explore machine reading comprehension from a more comprehensive perspective.

\begin{figure}
\centering
\includegraphics[width=0.5\textwidth,height=0.35\textwidth]{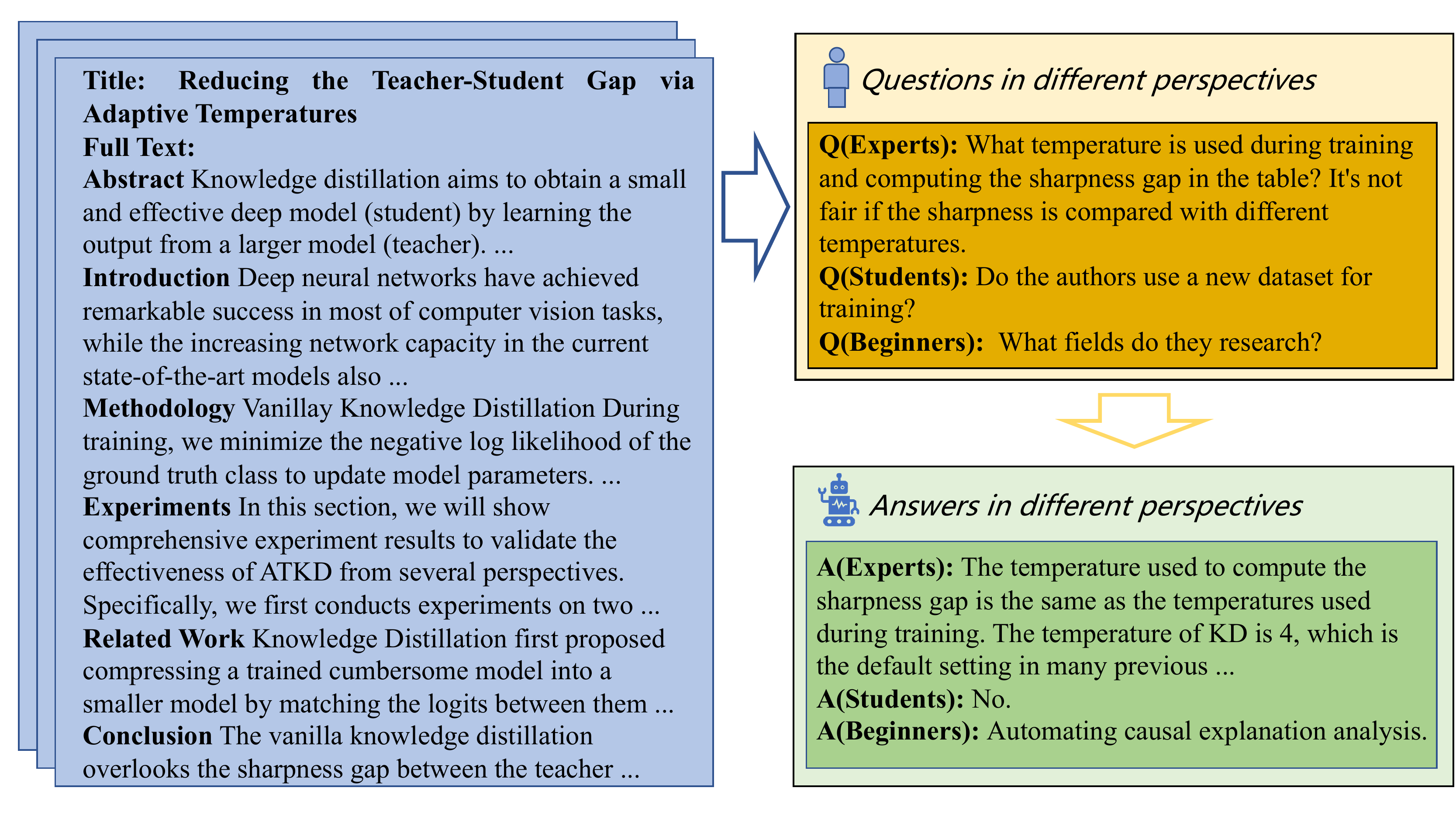}
\caption{An instance of \ours{}, given papers the questions from three perspectives raised by different users are considered, and machines are required to give corresponding answers to meet the needs of users with different levels of understanding.}
\label{fig:dis}
\end{figure}

\begin{table*}
\setlength{\tabcolsep}{3pt}
\centering\centering
\scalebox{0.85}{
\begin{tabular}{l c c c c c c c c c}
\toprule
\hline

Type & Paper & Figure/Table &
Question & \multicolumn{5}{c}{Answer} & Evidence\\
  \hline 
\textsc{Perspective}
 & \makecell[c]{Avg \\ Paper \\ Length} &\makecell[c]{Avg \\  Figure/Table \\ Number}  & \makecell[c]{Avg \\ Question \\ Length} & \makecell[c]{Avg \\ Answer \\ Length} & Yes|No  & Generative  & Extractive & Unanswerable & \makecell[c]{Avg \\ Evidence \\ Sentence \\ Number}
 \\ 
\hline
\textsc{Beginners}& \multirow{4}{*}{3725.6} & \multirow{4}{*}{5.32} & 10.0 & 17.2 & 331 &  754 & 2220 & 1 & 1.39  \\
\textsc{Students}&  &  & 9.8 & 11.7 & 266 & 340 &  1194 & 0 & 1.08  \\
\textsc{Experts}&  &  & 22.4 & 95.9 & 5 &  467 & 8 & 471 & 4.56\\
\textsc{All}& & & 11.0 & 21.8 & 602 & 1561 & 3422 & 472 & 1.56 \\
\hline
\bottomrule

\end{tabular}
}
\caption{Representative features from \ours{} categorized by different perspectives. }\label{table:dataset_analyse}
\end{table*}

To tackle the above problem. We propose \ours{}, a multi-perspective scientific paper machine reading comprehension dataset, with perspectives of the beginner's perspective, the student's perspective and the expert's perspective. The annotation tasks corresponding to these three perspectives are defined as: beginner perspective annotation by non-expert annotators, student perspective annotation guided with summarization knowledge, and expert perspective annotation based on open-review expert review. Each of the questions is classified into 28 categories. These 28 categories are obtained by analyzing and sorting out the issues of concern to the papers collected by relevant experts and students in colleges and universities in the early stage.

Our proposed \ours{} contains 741 papers and 6,057 QA pairs in total. Each perspective of beginners, students and experts contains 3,306, 1,800 and 951 QA pairs, respectively. Extensive experiments on \ours{} by utilizing pre-trained models with different perspectives settings are conducted to investigate the influence of perspectives for scientific machine reading comprehension, which suggests the importance of considering perspectives of SMRC, and demonstrates its challenging nature for machine comprehension. 

Our contributions are summarized as follows:
\begin{itemize}
    \item We propose a novel multi-perspective scientific machine reading comprehension dataset, called \ours{}, with three perspectives from beginners, students and experts.
    \item Extensive experiments on \ours{} by using pre-trained models suggest the importance of considering perspectives of SMRC, and demonstrate its challenging nature for machine comprehension. 
\end{itemize}

% We conduct the state-of-art methods to quantize the different perspectives in \ours{}. Experimental results suggest that xxx.

\section{\ours{}}
Considering different readers' perspectives for different levels of understanding of papers, we construct \ours{}, a multi-perspective science paper reading comprehension dataset. \ours{} contains 741 papers and 6,057 QA pairs. Besides, the question categories in \ours{} cover most issues of concern collected from relevant experts and students. In what follows, we describe how we construct the three perspectives of the \ours{} dataset.
\begin{figure}
\centering
\includegraphics[width=0.5\textwidth]{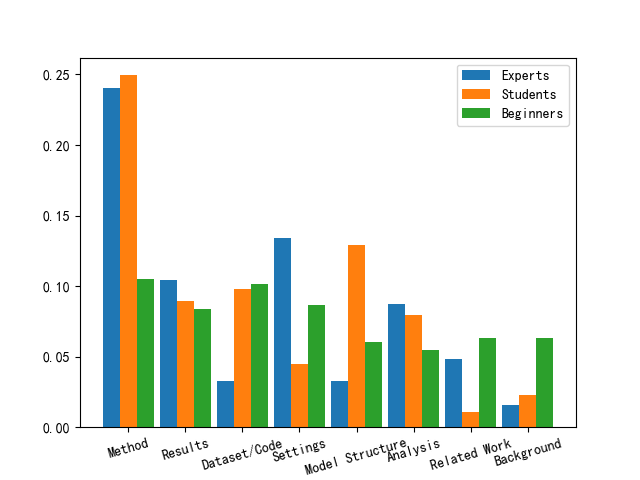}
\caption{The distribution of question categories}
\label{fig:dis}
\end{figure}

\subsection{Data Preparation}
We obtain relevant papers from s2orc \citep{Lo2020S2ORCTS} and open-review, and obtain pure text through the parser.
The total collected number of papers is 3k. To better obtain readers' interest in different contents of the scientific paper, we design a questionnaire in the early stage to collect the key points of experts and students in related fields when reading the paper, as well as the five most concerned issues. By summarizing the results of the questionnaire, we divide 780 readers' questions into 28 categories. These collected questions are used as reference questions for the annotators. After completing the preliminary data preparation, we will introduce the details of our data annotation from three different perspectives.

\paragraph{Questions Collection}
The annotators will read different materials according to the requirements of different perspectives, ask or extract questions, and map the questions to one of the 28 categories.

\begin{table*}
\setlength{\tabcolsep}{3pt}
\centering\centering
\scalebox{0.85}{
\begin{tabular}{l c c c c}
\toprule
\multicolumn{2}{l}{\textbf{Question}} & \textbf{Perspective} & \textbf{\%} \\
  \hline 
% \bottomrule
\multicolumn{2}{l}{\makecell[l]{How does the number of hierarchical levels pre-defined in real scenarios? The two \\experiments used L(m) = 2, will that be better to set it as a tuned parameter?}} & Experts & 15.7  \\ \hline 
\multicolumn{2}{l}{\makecell[l]{What are the advantages of their model compared with the existing MRC systems?}} & Students & 29.7  \\   \hline 
\multicolumn{2}{l}{How is the data for training collected?} & Beginners & 54.6  \\  
   
\bottomrule
\toprule
\textbf{Question} &  \textbf{Answer} & \textbf{Type} & \textbf{\%} \\ \hline
\makecell[l]{What task leaderboard does the paper describe?} & \makecell[l]{The commonsense reasoning task leaderboard \\ of the AI2 WinoGrande Challenge.} & Extractive & 56.5  \\ \hline 
\makecell[l]{What is the function of the modality attention \\ module in this paper?} & \makecell[l]{It selectively chooses modalities to extract \\ primary context from, maximizing information \\ gain and suppressing irrelevant contexts from \\ each modality.} & Generative & 25.7 \\ \hline 
\makecell[l]{Do the authors use a new dataset for training?} & No. & Yes|No & 10.0 \\ \hline 
\makecell[l]{Why is "sparse linear system[s]" preferred here \\ instead of general linear and nonlinear ones?} & N/A & Unanswerable & 7.8 \\
\bottomrule

\end{tabular}
}
\caption{Examples of questions, answers in different types sampled from \ours{}. \% are relative frequencies of the corresponding type over all examples in \ours{}}\label{table:dataset_case}
\end{table*}

\paragraph{Answers Collection}
The annotators will find the corresponding answers and evidence for the questions that have been asked according to the requirements of different perspectives.
\begin{itemize}
\item First of all, the annotator needs to judge whether the question can be answered. If no evidence of the question can be found in the paper, then the question is an \textsc{Unanswerable} question.
\item If the question can be answered, it is necessary to mark the corresponding evidence and map the answer to \textsc{Extraction}, \textsc{Generative}, or \textsc{Yes/no} type.
\end{itemize}

\subsection{Multi-perspective Collection}

We divide readers' perspectives into beginner, student and expert perspectives to better model the needs of different readers for scientific reading comprehension. Our tasks are completed by a professional labeling company and quality-checked by graduate students in relevant area. 

\paragraph{Beginner's perspective}
Beginners often don't have much experience of reading papers, and don't know much about the field. We ask annotators with no domain experience to ask questions about the paper and find the corresponding answers in the full text, figures and tables in the paper.

\paragraph{Student's perspective}
From the perspective of students, they often have a certain preliminary understanding of the academic field and have their views on the purpose, method, finding and value of the paper. In order to model this way of understanding, we use the summarization dataset FacetSum \cite{meng-etal-2021-bringing} to train the model and simplify the full text of a paper into four aspects: purpose, method, finding and value. The annotators were asked to first judge whether the generated abstracts were correct, and then were asked to ask questions by reading the correct abstracts, and furthermore to find the answers in the full text, figures and tables.

\paragraph{Expert's perspective}
Experts often have their own unique insights about this field and are able to give detailed opinions and development insights to papers. We crawled related reviews in open-review. The annotator extracts the corresponding questions and answers through the reviewer's review and the author's reply. In addition, the expert's questions require the background knowledge which is beyond the paper itself, and we set it as an \textsc{Unanswerable} question for this type of question. 
The model's ability to identify and answer questions beyond the scope of knowledge can be measured through the \textsc{Unanswerable} questions. 
% We can measure the model's ability to identify questions beyond the scope of knowledge, and the model's performance on answers beyond the scope of knowledge.

\section{Analysis About \ours{}}
\ours{} includes 741 papers and 6,057 QA pairs, which are divided into 3,306, 1,800 and 951 QA pairs corresponding to beginner's perspective, student's perspective and expert's perspective. 
We split our dataset into train, validation and test sets following 7:1:2 randomly.
Table \ref{table:dataset_analyse} provides representative features from \ours{} categorized by different perspectives. In the following, we describe it in detail.

\subsection{Question types}
Recall that we summarize 28 question categories from 780 questions which are specialized in scientific papers and cover different perspectives with different levels of understanding of scientific papers.
Besides, we also require the annotators to annotate questions and answers concerning the above question categories.
It helps us analyze the different concerns of different perspectives. Figure \ref{fig:dis} shows the distribution of question categories in the dataset under different perspectives. 
There are obvious differences in the question distribution under different perspectives.
Compared with the other two types of perspectives, beginners have a larger proportion of questions about the background knowledge of the scientific paper that experts and students know. Besides, experts and students also pay different attention to scientific papers. Students usually focus more on the code and model architecture, while experts focus more on the analysis of experiments. 
% In addition, to get a better sense of question types in the dataset, we sample several question examples from different perspectives in Table \ref{table:dataset_case}.

\subsection{Evidence Selection}
The evidence that annotators selected is also required to be concise. Therefore, the comprehensiveness and difficulty of questions can be implicitly reflected by the number of sentences of annotated evidence. Table \ref{table:dataset_analyse} shows the average number of sentences\footnote{we use "." to divide sentences and filter out any "." in reference.} of evidence in different perspectives. 
We find the sentence number of evidence in the expert's perspective is much larger than the other two perspectives, indicating the expert's questions are more comprehensive and harder.

\subsection{Answer types}
As shown in Table \ref{table:dataset_analyse}, three different perspectives have their distribution of answer types. Students and beginners have a similar distribution. 
Considering experts may have more questions about experiments and motivation that are not mentioned in papers, the proportion of \textsc{Unanswerable} questions and generative answers thus increases compared to others. Besides, we also find that the length of the answer in expert perspective is much longer than other perspectives.

\begin{table*}
\setlength{\tabcolsep}{3pt}
\centering\centering
\scalebox{0.95}{
\begin{tabular}{lccccccccc}
\toprule
\multirow{2}{*}{Model} &
\multicolumn{4}{c}{Dev} & \multicolumn{4}{c}{Test} \\
 & Beginner  & Student  & Expert & Overall 
 & Beginner  & Student  & Expert & Overall
 \\ \midrule
T5-B & 24.77\scriptsize{$\pm0.27$} & 39.50\scriptsize{$\pm0.72$} & 11.29\scriptsize{$\pm0.45$} & 26.47\scriptsize{$\pm0.23$} & 25.60\scriptsize{$\pm0.34$} & 41.51\scriptsize{$\pm0.44$} & 11.39\scriptsize{$\pm0.20$} & 29.03\scriptsize{$\pm0.13$} \\ 
T5-S & 24.32\scriptsize{$\pm0.11$} & 45.08\scriptsize{$\pm0.98$} & 10.64\scriptsize{$\pm0.28$} & 27.80\scriptsize{$\pm0.40$} & 24.68\scriptsize{$\pm0.31$} & 44.16\scriptsize{$\pm0.59$} & 9.97\scriptsize{$\pm0.13$} & 29.16\scriptsize{$\pm0.33$} \\
T5-E & 9.74\scriptsize{$\pm0.16$} & 17.03\scriptsize{$\pm0.57$} & \textbf{14.76}\scriptsize{$\pm0.20$} & 13.11\scriptsize{$\pm0.18$} & 11.12\scriptsize{$\pm0.21$} & 18.19\scriptsize{$\pm0.66$} & \textbf{15.66}\scriptsize{$\pm0.19$} & 13.79\scriptsize{$\pm0.34$} \\
T5-SE & 23.50\scriptsize{$\pm0.33$} & 45.17\scriptsize{$\pm0.28$} & 13.34\scriptsize{$\pm0.07$} & 28.09\scriptsize{$\pm0.08$} & 24.26\scriptsize{$\pm0.29$} & 43.34\scriptsize{$\pm0.84$} & 13.97\scriptsize{$\pm0.43$} & 29.12\scriptsize{$\pm0.21$} \\
T5-BE & 25.18\scriptsize{$\pm0.49$} & 39.65\scriptsize{$\pm0.47$} & 13.64\scriptsize{$\pm0.71$} & 27.21\scriptsize{$\pm0.46$} & 25.94\scriptsize{$\pm0.02$} & 41.50\scriptsize{$\pm0.82$} & 14.68\scriptsize{$\pm0.61$} & 29.61\scriptsize{$\pm0.30$} \\
T5-SB & 26.06\scriptsize{$\pm0.54$} & 45.93\scriptsize{$\pm0.87$} & 11.24\scriptsize{$\pm0.16$} & 29.12\scriptsize{$\pm0.41$} & \textbf{26.46}\scriptsize{$\pm0.15$} & \textbf{46.27}\scriptsize{$\pm0.22$} & 10.92\scriptsize{$\pm0.09$} & 30.99\scriptsize{$\pm0.21$} \\
T5-BSE & \textbf{26.87}\scriptsize{$\pm0.89$} & \textbf{46.86}\scriptsize{$\pm0.72$} & 13.60\scriptsize{$\pm0.24$} & 30.25\scriptsize{$\pm0.70$} & 26.34\scriptsize{$\pm0.49$} & 45.28\scriptsize{$\pm0.24$} & 14.53\scriptsize{$\pm0.28$} & 31.05\scriptsize{$\pm0.23$} \\
PerT5 & \textbf{26.87}\scriptsize{$\pm0.89$} & \textbf{46.86}\scriptsize{$\pm0.72$} & \textbf{14.76}\scriptsize{$\pm0.20$} & \textbf{30.45}\scriptsize{$\pm0.60$} & 26.34\scriptsize{$\pm0.49$} & 45.28\scriptsize{$\pm0.24$} & \textbf{15.66}\scriptsize{$\pm0.19$} & \textbf{31.18}\scriptsize{$\pm0.23$} \\

\midrule
LED-B & 25.51\scriptsize{$\pm1.61$} & 32.1\scriptsize{$\pm2.06$} & 10.02\scriptsize{$\pm0.29$} & 24.25\scriptsize{$\pm1.33$} & 25.02\scriptsize{$\pm1.21$} & 33.15\scriptsize{$\pm0.67$} & 11.36\scriptsize{$\pm0.23$} & 26.15\scriptsize{$\pm0.93$} \\
LED-S & 22.55\scriptsize{$\pm2.26$} & 41.93\scriptsize{$\pm6.73$} & 9.16\scriptsize{$\pm0.5$} & 25.77\scriptsize{$\pm3.11$} & 21.64\scriptsize{$\pm1.95$} & 44.05\scriptsize{$\pm5.12$} & 9.69\scriptsize{$\pm0.47$} & 27.47\scriptsize{$\pm2.72$} \\
LED-E & 6.80\scriptsize{$\pm0.49$} & 7.75\scriptsize{$\pm0.47$} & 14.62\scriptsize{$\pm0.32$} & 8.76\scriptsize{$\pm0.45$} & 7.22\scriptsize{$\pm0.53$} & 7.98\scriptsize{$\pm0.75$} & \textbf{14.77}\scriptsize{$\pm0.11$} & 8.27\scriptsize{$\pm0.53$} \\
LED-SE & 24.57\scriptsize{$\pm0.35$} & 45.16\scriptsize{$\pm1.03$} & 13.29\scriptsize{$\pm0.34$} & 28.55\scriptsize{$\pm0.48$} & 23.79\scriptsize{$\pm0.27$} & 47.42\scriptsize{$\pm1.23$} & 12.52\scriptsize{$\pm0.48$} & 30.00\scriptsize{$\pm0.14$} \\
LED-BE & 27.40\scriptsize{$\pm0.94$} & 32.62\scriptsize{$\pm1.15$} & 13.02\scriptsize{$\pm0.69$} & 25.89\scriptsize{$\pm0.64$} & 25.80\scriptsize{$\pm0.95$} & 33.81\scriptsize{$\pm1.03$} & 12.68\scriptsize{$\pm0.15$} & 26.90\scriptsize{$\pm0.88$} \\
LED-SB & 31.07\scriptsize{$\pm0.07$} & \textbf{49.67}\scriptsize{$\pm0.27$} & 9.07\scriptsize{$\pm0.3$} & 32.17\scriptsize{$\pm0.25$} & 29.78\scriptsize{$\pm0.4$} & \textbf{47.93}\scriptsize{$\pm0.42$} & 10.41\scriptsize{$\pm0.39$} & 33.37\scriptsize{$\pm0.41$} \\
LED-BSE & \textbf{32.24}\scriptsize{$\pm0.43$} & 47.04\scriptsize{$\pm2.38$} & 12.85\scriptsize{$\pm0.75$} & 32.04\scriptsize{$\pm0.87$} & \textbf{29.80}\scriptsize{$\pm0.55$} & 46.39\scriptsize{$\pm0.34$} & 12.54\scriptsize{$\pm0.08$} & 33.20\scriptsize{$\pm0.36$} \\
PerLED & \textbf{32.24}\scriptsize{$\pm0.43$} & 47.04\scriptsize{$\pm2.38$} & \textbf{14.62}\scriptsize{$\pm0.32$} & \textbf{32.47}\scriptsize{$\pm0.69$} & \textbf{29.80}\scriptsize{$\pm0.55$} & 46.39\scriptsize{$\pm0.34$} & \textbf{14.77}\scriptsize{$\pm0.11$} & \textbf{33.44}\scriptsize{$\pm0.34$} \\

\bottomrule
\end{tabular}
}
\caption{The performance of answer generation on the validation and test set with different eval settings. The average results with standard deviation on 3 random seeds are reported.}\label{table: dev-test-subEval}
\end{table*}
\section{Modeling}
To unify the different answer types of \ours{}, we conduct text-to-text transformers, including T5 \cite{t5} and LED \cite{beltagy2020longformer}, to model the scientific machine reading comprehension task. 

% Pre-trained Transformer \citep{NIPSattention} model becomes popular in QA tasks, since its outstanding performance. 
% Since the answer types in \ours{} are diverse, 
% However, the answer types in \ours{} include \textsc{Extractive}, \textsc{Generative}, \textsc{Yes|No} and \textsc{Unanswerable} question. It's so diverse that original Bert-based QA model \cite{devlin-etal-2019-bert} is not sufficient to cover all the answer types. 
% Therefore,
% we treat all the answer types as a "text-to-text" problem, which is inspired by \citep{t5} and \citep{dasigi2021dataset}. Besides, long input document is still a challenge for typical Transformer encoder-decoder architecture. We utilize LED \cite{beltagy2020longformer} that replace complex self-attention with efficient sliding window attention patterns to alleviate the performance gap caused by token limitation.

\paragraph{Text-to-Text Transformer}
Text-to-Text Transformer (T5) \cite{t5} is based on Transformer\citep{} encoder-decoder architecture. It treats every text processing problem such as question answering and classification as a "text-to-text" problem, each fine-tuning task is specifically prefixed with a task prefix.

\paragraph{Longformer Encoder Decoder}
Longformer Encoder Decoder (LED)  is a special Transformer encoder-decoder model which combines local and global attention patterns. It replaces original self-attention with several sliding window attention patterns. Compared to full text self-attention with quadratic computation, LED scales its self-attention computation linearly with the input size. Besides, the limitation of input size is also up to 16k tokens (as opposed to 1k tokens in the original T5 model).

% It allows LED to calculate only some of tokens rather than all of them in the local window. Therefore, 

% \subsection{Task setup}

% \subsection{Human performance}
\subsection{Training objective}
Given the context, question and the final answer $a$, the answer is composed of the variable-length tokens $x_{i}$, and the probabilities over the tokens are shown as following:
    \begin{equation}
        p(a) = \prod_{1}^{m}p(x_{i}|x_{<i},f_{e};\theta ),
    \end{equation}
where $\theta$ donates the trainable parameters of our model. The training objective is computed as illustrated as following:
    \begin{equation}
        \mathcal{L}_{oss} = -\sum_{i=1}^{M} log \, p(x_{i}|x_{<i},f_{w};\theta ),
    \end{equation}
\subsection{Metrics}
Considering there is \textsc{Yes|No} question type in \ours{}, in addition to Rouge-L, we also use accuracy as an automatic proxy for the measure of correctness of answers.
\paragraph{Rouge-L}Since there are many generative answers in our datasets, we evaluate our reading comprehension task via Rouge-L \citep{Lin2004ROUGEAP} \footnote{The implementation we used is from huggingface (https://huggingface.co/)}, which is widely used in language generation evaluation. It utilizes the length of the longest common sequence (LCS), predicted answer (Prediction) and golden answer (Golden) to calculate the final score as follows:
    \begin{equation}
        \mathcal{R}_{LCS} = \frac{LCS(Prediction, Golden)}{len(Golden)}
    \end{equation}
    \begin{equation}
        \mathcal{P}_{LCS} = \frac{LCS(Prediction, Golden)}{len(Prediction)}
    \end{equation}
    \begin{equation}
        \mathcal{F}_{LCS} = \frac{(1+\beta^2)\mathcal{R}_{LCS}\mathcal{P}_{LCS}}{\mathcal{R}_{LCS} + \beta^2\mathcal{P}_{LCS}}
    \end{equation}

\paragraph{Accuracy}
For \textsc{Yes|No} questions, accuracy is utilized for evaluation. For "yes" questions, we recognize an answer that starts with "yes" as a correct answer. For the "no" questions, we treat answers that do not start with "yes" as correct answers.

\begin{table*}
\setlength{\tabcolsep}{3pt}
\centering\centering
\scalebox{0.9}{
\begin{tabular}{lcccccccc}
\toprule
\multirow{2}{*}{Model} &
\multicolumn{4}{c}{Dev} & \multicolumn{4}{c}{Test} \\
 & Yes|No  & Generative  & Extractive & Unanswerable  &
 Yes|No  & Generative  & Extractive & Unanswerable 
 \\ \midrule
T5-BSE & 81.61\scriptsize{$\pm1.63$} & 25.67\scriptsize{$\pm0.12$} & 30.27\scriptsize{$\pm0.84$} & 13.04\scriptsize{$\pm0.33$} & \textbf{83.61}\scriptsize{$\pm0.67$} & 23.16\scriptsize{$\pm0.11$} & 29.58\scriptsize{$\pm0.24$} & 14.86\scriptsize{$\pm0.30$} \\
PerT5 & 81.61\scriptsize{$\pm1.63$} & 26.14\scriptsize{$\pm0.30$} & 30.38\scriptsize{$\pm0.93$} & 14.23\scriptsize{$\pm0.22$} & \textbf{83.61}\scriptsize{$\pm0.67$} & 23.39\scriptsize{$\pm0.07$} & 29.61\scriptsize{$\pm0.21$} & \textbf{16.11}\scriptsize{$\pm0.03$} \\
LED-BSE & \textbf{82.18}\scriptsize{$\pm0.82$} & 25.56\scriptsize{$\pm0.81$} & 34.78\scriptsize{$\pm1.43$} & 12.79\scriptsize{$\pm0.57$} & 78.69\scriptsize{$\pm1.34$} & 24.81\scriptsize{$\pm0.36$} & 33.54\scriptsize{$\pm0.36$} & 13.17\scriptsize{$\pm0.29$} \\
PerLED & \textbf{82.18}\scriptsize{$\pm0.82$} & \textbf{26.32}\scriptsize{$\pm0.79$} & \textbf{34.83}\scriptsize{$\pm1.44$} & \textbf{14.53}\scriptsize{$\pm0.11$} & 78.69\scriptsize{$\pm1.34$} & \textbf{25.66}\scriptsize{$\pm0.40$} & \textbf{33.57}\scriptsize{$\pm0.33$} & 13.99\scriptsize{$\pm0.23$} \\
\bottomrule
\end{tabular}
}
\caption{The performance of different answer types on the validation set and test set. The
average results with standard deviation on 3 random seeds are reported.}\label{table: dev-test}
\end{table*}

\section{Experiments}

 The experimental results are illustrated in Table \ref{table: dev-test-subEval}, which demonstrates the challenges of the existing model facing multi-perspective problems in \ours{}. To investigate the influence of multi-perspective in \ours{}, T5 and LED are chosen as our backbone model with different training settings by combining different perspectives of beginners, students and experts.  In addition, we conduct experiments to investigate performance of the different answer formats in \ours{} on \textsc{Yes|No}, \textsc{Generative}, \textsc{Extractive} and \textsc{Unanswerable} subset, which is shown in Table \ref{table: dev-test}. There are several  experimental settings are considered: 
\begin{itemize}

\item \textbf{B-formed} Beginner's perspective training data is used to train the model. 
\item \textbf{S-formed} Student's perspective training data is used to train the model.
\item \textbf{E-formed} Expert's perspective training data is used to train the model.
 % \item \textbf{S}-formed trains the model by using student's perspective training data. 
 % \item \textbf{E}-formed trains the model by using expert's perspective training data.
\item \textbf{BS-formed} Beginner's and student's perspective training data is used to train the model. 
\item \textbf{BE-formed} Beginner's and expert's perspective training data is used to train the model.
\item \textbf{SE-formed} Student's and expert's perspective training data is used to train the model.
% \item \textbf{BS}-formed trains the model by combining beginner's and student's perspectives training data.
% \item \textbf{BE}-formed trains the model by combining beginner's and expert's perspectives training data.
% \item \textbf{SE}-formed trains the model by combining student's and expert's perspectives training data.
\item \textbf{BSE-formed} All training data is used to train the model.
\item \textbf{Pre-formed} Pre-formed model consists of \textsc{BSE}-formed model and \textsc{BSE}-formed model, which is utilized in the known question perspectives settings. \textsc{E}-formed model is utilized for evaluating expert perspective, \textsc{BSE}-formed model is utilized for evaluating student and beginner perspectives.
% \begin{itemize}
% \item \textbf{-B} trains the model by using beginner perspective training data.
% \item \textbf{-S} trains the model by using student perspective training data.
% \item \textbf{-E} trains the model by using expert perspective training data.
\end{itemize}

\subsection{Results}
\begin{figure*}[t]
\centering
\includegraphics[width=0.99\textwidth,height=0.4\textwidth]{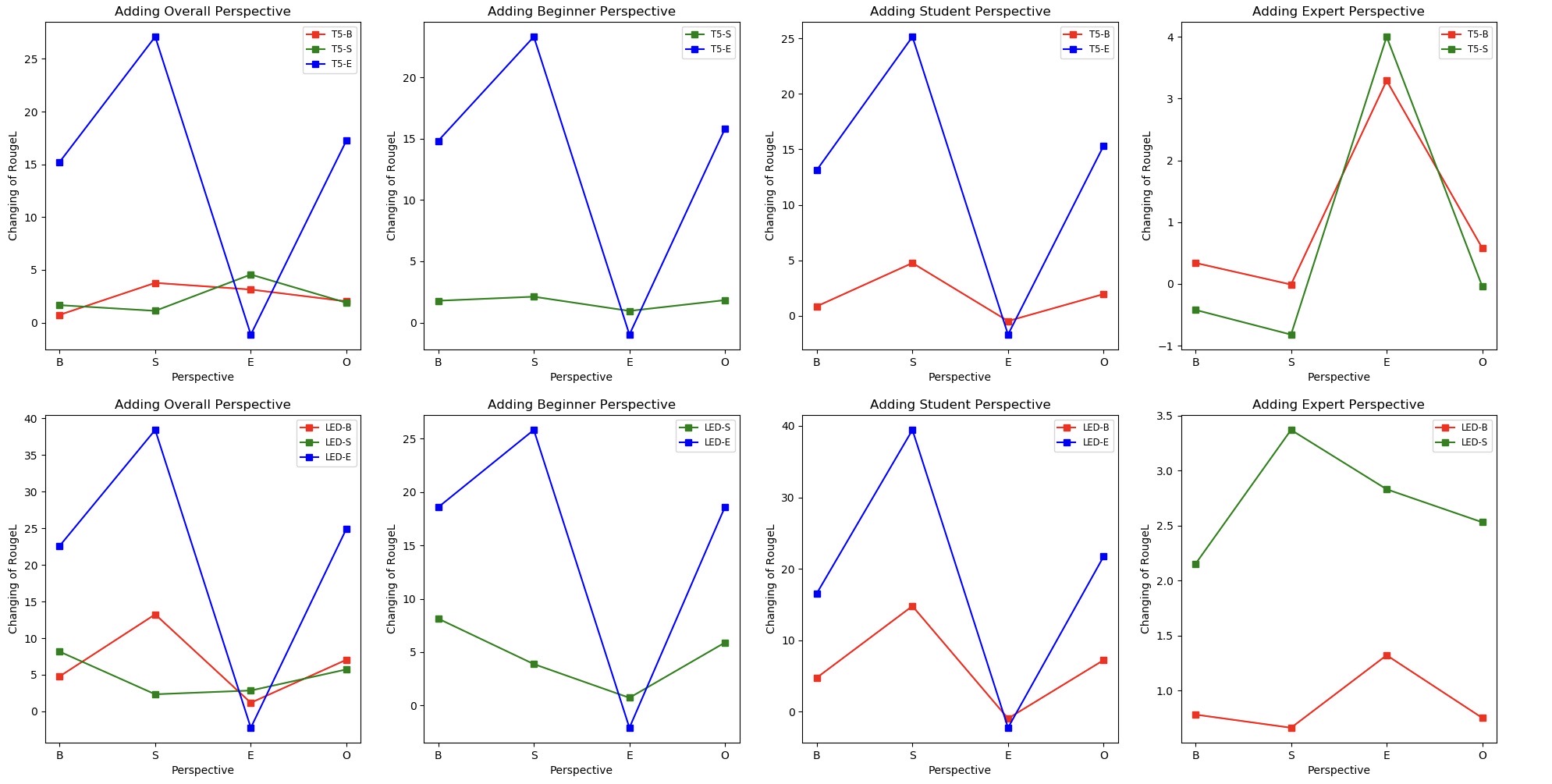}
\caption{Based on the data of either perspective of \textsc{S}, \textsc{B} and \textsc{E}, we further add other perspective data, and then observe the change of Rouge-L scores on test set to quantity the influence of performance.}
\label{fig:t5Perspective}
\end{figure*}
\paragraph{Multi-Perspective}
 To investigate the influence of multi-perspective in \ours{}, we user T5 and LED as our backbone model with different training settings and evaluated the performance of perspectives: beginner, student, expert and overall. The evaluation was conducted on both a development set and a test set, with Rouge-L scores separated by beginner, student, expert and overall perspectives. The results, presented in Table \ref{table: dev-test-subEval}, show that the perspective-formed models: \modelT, \modelLED, perform the best Rouge-L scores on the development set and test set. \modelLED \space achieves the highest performance through the assumption of known perspectives, with scores of 33.44 on the test set. \textsc{E}-formed models perform best for the expert perspective, with an overall mean performance of 15.66 and 14.77 on the test set, respectively. In general, the models have a higher performance on the student and beginner perspectives compared to the expert perspective, which demonstrates that the expert perspective is more difficult than the student and beginner perspectives on \ours{}.

 % As illustrated in Table \ref{table: dev-test-subEval}, the extensive experimental results suggest the importance of multi-perspective. Different perspective influence other perspectives, resulting in the different understanding levels of texts, which demonstrates the challenges of existing model facing multi-perspective problems in \ours{}.
 
 % In addition, we conduct experiment to investigate influence of the different answer format in \ours{} by removing either of Yes|No, Generative, Extractive, \textsc{Unanswerable} subset, which is 
 \paragraph{Multi-Type of Answer}
 Table \ref{table: dev-test} presents the results of an experiment comparing several models (T5-BSE, \modelT, LED-BSE, \modelLED) on \ours{}. The performance is measured in terms of four categories of answer types: \textsc{Yes|No}, \textsc{Generative}, \textsc{Extractive} and \textsc{\textsc{Unanswerable}}. The results are presented for both the dev and test sets. Comparing the performance of the models on the \textsc{Yes|No} answer type, T5-BSE, \modelT \space achieves the highest performance on the test set with scores of 83.61 and 83.61, respectively. On the \textsc{Generative} answer type, \modelT \space performed worse than \modelLED \space on both the dev set and test set, with a score of  23.39 and 25.66 respectively. On the \textsc{Extractive} answer type, \modelLED \space achieves the best performance on the test set with a score of 33.57. On the \textsc{\textsc{Unanswerable}} answer type, \modelT \space achieves the best performance on the test set with a score of 16.11.

\subsection{Analysis}

\paragraph{Analysis of Perspective and Performance}

As shown in Table \ref{table: dev-test-subEval}, it can be observed the performance of the models is closely related to the perspectives, thus we further investigate the influence of performance by analyzing the different perspective settings on the test set. The results show that adding the beginner perspective to the model improves the overall performance, as seen by the higher performance of \textsc{SB}-formed and \textsc{BSE}-formed as compared to \textsc{S}-formed and \textsc{SE}-formed. Similarly, it also has a positive impact on the overall performance by adding the student perspective, as seen by the higher performance of \textsc{B}-formed and \textsc{BE}-formed. The results also suggest that the expert perspective has a positive impact on the overall performance as well. The T5-E model, which only includes the expert perspective, has the highest mean performance for the expert perspective with a score of 15.66 on the test set. Additionally, \textsc{SE}-formed, \textsc{BE}-formed, \textsc{SB}-formed and \textsc{BSE}-formed which includes the expert perspective, all have higher overall mean performance compared to \textsc{S}-formed, \textsc{B}-formed and \textsc{SB}-formed which does not include the expert perspective. 

% The above results demonstrate the importance of analyzing the perspectives of scientific machine reading comprehension. 

\paragraph{Analysis of Characteristic of Perspectives}

As shown in Figure \ref{fig:t5Perspective}, the inner relations among perspectives are illustrated, where the fusion of the perspectives of the student and the beginner can improve the understanding ability of the model in each perspective, further adding additional data from the expert perspective will enhance the understanding ability of the expert perspective, but it will slightly reduce the performance of other perspectives. Moreover, the models have higher performance on the student and beginner perspectives compared to the expert perspective, which demonstrates that the expert perspective is more difficult than the student and beginner perspectives on \ours{}. This can be seen from the lower performance scores of the models in the expert perspective compared to the student and beginner perspectives. The expert perspective requires a higher level of understanding and knowledge of the subject matter because the questions were asked by experienced reviewers who have a deeper understanding of the topic, which requires a more complex understanding and more external knowledge of the subject matter compared to the student and beginner perspectives. This would also make the expert perspective more challenging for the models to predict. The above results suggest the insights and challenges of the current pre-trained models on  \ours{}. 
% It is important to note that the expert perspective also include questions that

\begin{table*}
\setlength{\tabcolsep}{3pt}
\centering\centering
\scalebox{0.6}{
\begin{tabular}{ p{3cm}  p{5cm}  p{5cm}  p{10cm}  p{2cm} }

\toprule
\textbf{Title} &\textbf{Question} &  \textbf{Prediction} &\textbf{Golden} & \textbf{Perspective}  \\ \hline
Stochastic Answer Networks for SQuAD 2.0 & What data sets did the author use? & SQuAD 2.0 dataset, a new MRC dataset which is a combination of Stanford Question Answering Dataset (SQuAD) 1.0 and additional \textsc{Unanswerable} question-answer pairs. & They evaluate their system on SQuAD 2.0 dataset, a new MRC dataset which is a combination of Stanford Question Answering Dataset (SQuAD) 1.0 and additional \textsc{Unanswerable} question-answer pairs. & Beginner  \\  \hline 

From Textual Information Sources to Linked Data in the Agatha Project & What other related research path do the authors have explored? & The authors have already explored other related research path, namely author profiling BIBREF2, aggression identification BIBREF3 and hate-speech detection BIBREF4 over social media, plus statute law retrieval and entailment for Japanese. & Author profiling, aggression identification and hate-speech detection over social media, plus statute law retrieval and entailment for Japanese. & Student \\ \hline 
% F-Score Driven Max Margin  Neural Network for Named Entity Recognition  in Chinese Social Media   & What does the  semi-supervised model base on? & They construct a semi-supervised model based on B-LSTM neural network to  learn from the limited labelled corpus  by using lexical information  provided by massive unlabeled text.& B-LSTM neural network. & Student \\ \hline 
 Edge Partition Modulated Graph Convolutional Networks & As the authors imply, it is becoming more important to take into account real-world limitations when deploying MARL (e.g., communication failure). Could you discuss some relevance to a 2019 ICLR paper called SchedNet by Kim et al.? (which, to the best of my knowledge, is one of the earliest MARL works addressing real-life communications constraints. 
& We agree with Kim et al. that it is becoming more important to take into account real-world limitations when deploying MARL (e.g., communication failure). Therefore, we proposed the imputation assisted multiagent reinforcement learning (IA-MARL) that can achieve the performance of MARL without missing training data when sufficient imputation accuracy is supported.

& In Kim et al (2019), the authors proposed SchedNet for the environment having limited communication channels between agents “during execution”, but assumed that the training data always exists during training. Note that the communication between the agents during the execution is to encourage coordination among the agents. On the other hand, in our work, the training data collected from agents can be missed, which affects “training of agents”. To summarize, there are two differences between SchedNet and IA-MARL: 1) SchedNet considers limited communication during execution and IA-MARL considers data missing for the training 2) the number of messages in ScheNet is limited while IA-MARL cannot obtain the missed data. We will cite and discuss it in the related works.
 & Expert \\ \hline

% Efficient and Modular Implicit Differentiation & What is "Implicit Differentiation"? & The term "Implicit Differentiation"  is used to refer to the fact  that implicit differentiation can be used  to solve complex problems  such as hyper-parameter optimization,  neural networks, and meta-learning.  For example, in the case of  hyper-parameter optimization,  we can solve the  normal equation AAu = Av.& To make it clearer,  we rephrased  the introduction as follows: 
% In recent years, two main approaches  have been developed to  circumvent this problem.  
% The first one consists of  unrolling the iterations of  an optimization algorithm  and using the final iteration  as a proxy for the optimization problem solution. This allows to explicitly construct a computational graph relating the algorithm output to the inputs, on which autodiff can then be used transparently. In contrast, a second approach consists in implicitly relating an optimization problem solution to its inputs using optimality conditions.  In a machine learning context,  such implicit differentiation has been used for stationarity conditions, KKT conditions, and the proximal gradient fixed point.
%  & Expert \\ \hline 
\bottomrule

\end{tabular}
}
\caption{The predictions of \modelT \space and \modelLED \space on the test set of \ours{}.}\label{table:dataset_case}
\end{table*}

\paragraph{Analysis of Answer Types}

As shown in Table \ref{table: dev-test}, the scores for \textsc{\textsc{Unanswerable}} type are significantly lower than the scores for the other answer types. This is because the answers of \textsc{\textsc{Unanswerable}} all need external knowledge of specific academic topic and the answers cant directly reason through the context. Additionally, the LED-based models perform better than T5-based models for the answers of \textsc{Extractive} type, with scores of 33.57 for \modelLED, indicating the importance of the context length, where LED-based models utilize 2k tokens context and T5-based models only use 512 tokens context. Experimental results also suggest that current models facing challenges on \textsc{Extractive}, \textsc{Generative} and \textsc{\textsc{Unanswerable}} answer types, especially on the \textsc{\textsc{Unanswerable}} answer types, demonstrating the lack the inner academic knowledge of current pre-trained models. The above results demonstrate the challenges of the current pre-trained models on \ours{}.

\subsection{Case Study}
The predictions of \modelT \space and \modelLED \space are illustrated in Table \ref{table:dataset_case}. The student and beginner perspectives data are predicted by \modelLED, as it achieves the highest performance on the test set of student and beginner perspectives. The expert perspectives data are predicted by \modelT, as it achieves the highest performance on the test set of expert perspectives. As shown in Table \ref{table:dataset_case}, the predictions of student and beginner perspectives suggest the good understanding ability of models, and the predictions of expert perspective suggest that the understanding ability requirement of experts is more complicated than the requirement of beginners and students. Despite \modelT \space showing the potential of comprehending the context with the understanding level of experts, the performance is still limited. Reasoning through QA pairs of expert perspectives, machines not only need research background knowledge but also a deep understanding of the research area.

\section{Related work}

\paragraph{Information Seeking MRC Datasets}
MRC tasks can be regarded as information-seeking work, especially when reasoning from a question to an answer needs a complex information-seeking strategy. For example, SQuAD \citep{Rajpurkar2016SQuAD1Q}, TriviaQA \citep{joshi2017triviaqa} and HotpotQA \citep{yang2018hotpotqa} need to seek important information related to a question.  However, these datasets are mainly extractive, which constrains the flexibility and diversity of the QA pairs. Like \Qa the type of answers in \ours{} is diverse. It makes \ours{} closer to real-world settings. Furthermore, the questions of general domain MRC datasets including WikiQA \citep{Yang2015WikiQAAC}, Natural Questions \citep{Kwiatkowski2019NaturalQA} and IIRC \citep{Ferguson2020IIRCAD} are mainly based on common sense and context.  However, in these studies, there is a lack of discussion on the different levels of comprehension corresponding to different readers' perspectives. 

\paragraph{MRC Datasets in Academic Domain}
There are some MRC datasets targeting academic domains, where domain-specific knowledge is critical to tackling these problems. PubmedQA \citep{jin2019pubmedqa} is a biomedical MRC dataset, where the context is abstract, the question is the corresponding title and the answer can only be \textsc{Yes|No|Maybe}. BioMRC \citep{pappas2020biomrc} focuses on cloze-style MRC tasks in the biomedical domain. But these studies only conduct title and abstract as the context. As far as we know, there is only one study \cite{dasigi2021dataset} focused on full-text scientific machine reading comprehension. \Qa \cite{dasigi2021dataset} takes an entire paper as the context to do question answering, where the answer can be in various forms, including \textsc{Extractive}, \textsc{Abstractive}, \textsc{Yes|No} and \textsc{Unanswerable}. However, \Qa \cite{dasigi2021dataset} mainly takes the annotators as the single source of supervision, while multiple perspectives of annotations among different levels of researchers are needed in academic research works. \ours{} includes three perspectives of paper reading comprehension: the beginner's perspective, the student's perspective and the expert's perspective. 

% Besides, we summarize 28 question categories to ensure that \ours{} covers most of the concern of experts and students in relevant field.

% \ours{} also requires background knowledge in academic domain that varies from person to person.

% Afterwards, complex reasoning is applied to these information so as to produce the answer.

% So they focus more on the understanding of context.
% \paragraph{MRC Datasets in Multimodal}
% \paragraph{Long QA}

\section{Conclusion}
In this paper, we proposed a novel multi-perspective scientific machine (SMRC) reading comprehension dataset, called \ours{}, with different perspectives of readers, including beginners, students and experts. Extensive experimental results suggest the inner relations and differences among different perspectives, suggesting the importance of analyzing perspectives. The extensive results suggest that \ours{} could serve as a test-bed for evaluating SMRC research. 

% In the future, we will explore future solutions for multi-perspective understanding of machine reading comprehension based on perspective entry points.

% Expert perspective data should be specifically considered, due to its difficulty, which requires complicated understanding and external knowledge. 
\section*{Ethical Considerations}

We present a novel dataset that uses papers authored
by other researchers. \ours{} is an English MRC dataset targeting academic domains. To adhere to copyright, we
have restricted ourselves to arXiv papers released
under a CC-BY-* license. 
The payment for the annotators is well-above the minimum wage in our local area, and there is not any personal information in our dataset.

\section*{Limitations}
Our current multi-perspective research on \ours{} is limited to three perspectives: beginner, student and expert. In real life, the perspective of understanding scientific papers can be further generalized, even if it is an expert perspective, it can also be divided into senior and junior experts. In the future, more refinement about the perspectives is needed.

 % Our current multi-perspective research on \ours{} only concerns multi-perspective in English-language scientific machine reading comprehension dataset, and the extensive experiments are conducted in English-language pre-trained models. Future studies can further explore the research on  multilingual multi-perspective dataset, and conduct experiments on multilingual pre-trained models.

% We present a new dataset that uses papers authored
% by other researchers. To adhere to copyright, we
% have restricted ourselves to arXiv papers released
% under a CC-BY-* license, as identified via Unpay-
% wall, which was used in the S2ORC (Lo et al.,
% 2020) dataset construction. Due to our choice to
% use arXiv as the source of papers, \Qa is al-
% most entirely an English-language dataset, and QA
% systems built on \Qa would not be expected
% to work well on non-English language research
% papers.
% We have determined the amount we paid the
% annotators to be well-above the minimum wage in
% our local area. While we do collect information
% about annotator background in NLP and familiarity
% with the papers they are annotating, we have not
% collected personal identifiable information without
% their permission except for payment purposes, and
% do not include any such information in the released
% dataset

\bibliographystyle{acl_natbib}
\bibliography{cc}

\newpage
\appendix

\section*{Appendix}

\section{Implementation Details}
We utilize the T5-base \cite{t5} , LED-base \cite{beltagy2020longformer} as our reader backbone, and use AdamW \cite{loshchilov2018fixing-adamw} to fine-tune our models. The learning rate is set as 2e-5 for LED-base, and is set as 1e-4 for T5. And we use linear warm up scheduler, with warmup ratio of 0.1. Our final configuration is 8 epochs, and batch size
of 16. All the experiments conduct on two RTX 3090 GPUs, and each experiment takes
30–60 minutes for T5-base, 1-2 hours for LED-base. We implement \modelT \space and \modelLED \space with the PyTorch\footnote{\url{https://github.com/pytorch/pytorch}} library and using pre-trained Transformers from the Hugging Face\footnote{\url{https://github.com/huggingface/transformers}} repositories. The above repositories provide the data, models and licenses. 
\begin{table*}
\setlength{\tabcolsep}{3pt}
\centering\centering
\scalebox{0.85}{
\begin{tabular}{c c c c c c c c}
\toprule
1 & Methods & 2 & Experimental results & 3 & Dataset/code & 4 & Experimental settings \\ \hline
5 & Model architecture & 6 & Experimental analysis & 7 & Related work & 8 & Background  \\ \hline 
9 & Baseline & 10 & Motivation & 11 & Contribution & 12 & Innovation point \\ \hline
13 & Future work & 14 & Research field & 15 &  Research background & 16 & Evaluation Metric  \\ \hline
17 & Insufficiency of previous work & 18 & Experimental conditions & 19 & Assumption & 20 & Model transferability \\ \hline
21 & Limitations & 22 & Innovation & 23 & Thesis Outline & 24 & Landing Application \\ \hline
25 & Case Study & 26 & Publication Details & 27 & Complexity & 28 & Whether to propose a new task \\ 

\bottomrule
\end{tabular}
}
\caption{All the 28 question categories in \ours{}}\label{table:question_categories}
\end{table*}

\section{Question Categories}
Table \ref{table:question_categories} shows all the 28 question categories in \ours{}.

\end{document}